\documentclass[11pt,a4paper]{article}
\usepackage[hyperref]{acl2019}
\usepackage{times}
\usepackage{latexsym}

\usepackage{url}

\usepackage{graphicx}
\usepackage{pgfplots}
\usepackage{multirow}
\usepackage{siunitx}
\usepackage{amsmath}
\usepackage{booktabs}

\usepackage{tablefootnote}

\definecolor{bblue}{HTML}{4F81BD}
\definecolor{rred}{HTML}{c4260b}
\definecolor{ggreen}{HTML}{098c1f}
\definecolor{ppurple}{HTML}{9F4C7C}
\definecolor{oorange}{HTML}{F79646}

\aclfinalcopy 
\def\aclpaperid{1271} 

\title{Revisiting Low-Resource Neural Machine Translation:\\ A Case Study}

\author{Rico Sennrich$^{1,2}$ \quad Biao Zhang$^1$ \bigskip\\
  $^1$School of Informatics, University of Edinburgh \\
  \texttt{rico.sennrich@ed.ac.uk, B.Zhang@ed.ac.uk} \medskip\\
    $^2$Institute of Computational Linguistics, University of Zurich}

\date{}

\begin{document}
\maketitle
\begin{abstract}

It has been shown that the performance of neural machine translation (NMT) drops starkly in low-resource conditions,
underperforming phrase-based statistical machine translation (PBSMT) and requiring large amounts of auxiliary data to achieve competitive results.
In this paper, we re-assess the validity of these results, arguing that they are the result of lack of system adaptation to low-resource settings.
We discuss some pitfalls to be aware of when training low-resource NMT systems, and recent techniques that have shown to be especially helpful in low-resource settings, resulting in a set of best practices for low-resource NMT.
In our experiments on German--English with different amounts of IWSLT14 training data, we show that, without the use of any auxiliary monolingual or multilingual data, an optimized NMT system can outperform PBSMT with far less data than previously claimed.
We also apply these techniques to a low-resource Korean--English dataset, surpassing previously reported results by 4 BLEU.
\end{abstract}

\section{Introduction}

While neural machine translation (NMT) has achieved impressive performance in high-resource data conditions, becoming dominant in the field \citep{DBLP:conf/nips/SutskeverVL14, DBLP:journals/corr/BahdanauCB14, DBLP:journals/corr/VaswaniSPUJGKP17},
recent research has argued that these models are highly data-inefficient, and underperform phrase-based statistical machine translation (PBSMT) or unsupervised methods in low-data conditions \citep{koehn-knowles:2017:NMT, lample2018phrase}.
In this paper, we re-assess the validity of these results, arguing that they are the result of lack of system adaptation to low-resource settings.
Our main contributions are as follows:

\begin{itemize}
\item we explore best practices for low-resource NMT, evaluating their importance with ablation studies.
\item we reproduce a comparison of NMT and PBSMT in different data conditions, showing that when following our best practices, NMT outperforms PBSMT with as little as \num{100000} words of parallel training data.
\end{itemize}

\section{Related Work}

\subsection{Low-Resource Translation Quality Compared Across Systems}

\begin{figure}
\centering
\includegraphics[page=4, trim=2cm 19cm 10.3cm 2cm, clip, width=0.4\textwidth]{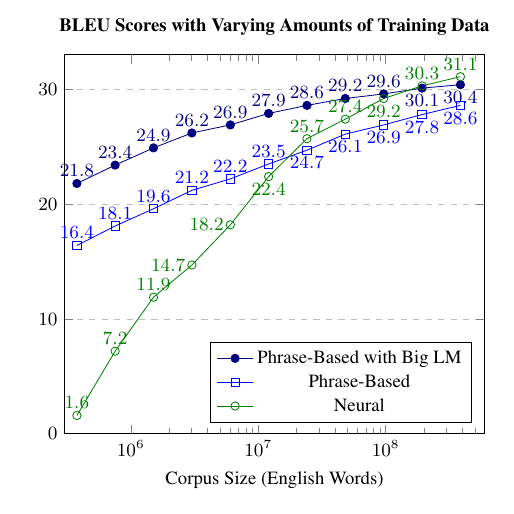}
\caption{quality of PBSMT and NMT in low-resource conditions according to \citep{koehn-knowles:2017:NMT}.}
\label{lrnmtplot-koehn}
\end{figure}

Figure \ref{lrnmtplot-koehn} reproduces a plot by \citet{koehn-knowles:2017:NMT} which shows that their NMT system only outperforms their PBSMT system when more than 100 million words (approx.\ 5 million sentences) of parallel training data are available.
Results shown by \citet{lample2018phrase} are similar, showing that unsupervised NMT outperforms supervised systems if few parallel resources are available.
In both papers, NMT systems are trained with hyperparameters that are typical for high-resource settings, and the authors did not tune hyperparameters, or change network architectures, to optimize NMT for low-resource conditions.

\subsection{Improving Low-Resource Neural Machine Translation}
\label{sec:transferlearning}

The bulk of research on low-resource NMT has focused on exploiting monolingual data, or parallel data involving other language pairs.
Methods to improve NMT with monolingual data range from the integration of a separately trained language model \citep{DBLP:journals/corr/GulcehreFXCBLBS15}
to the training of parts of the NMT model with additional objectives, including a language modelling objective \citep{DBLP:journals/corr/GulcehreFXCBLBS15,2015arXiv151106709S,ramachandran2016unsupervised},
an autoencoding objective \citep{DBLP:journals/corr/LuongLSVK15,currey-micelibarone-heafield:2017:WMT},
or a round-trip objective, where the model is trained to predict monolingual (target-side) training data that has been back-translated into the source language \citep{2015arXiv151106709S,NIPS2016_6469,cheng-EtAl:2016:P16-1}.
As an extreme case, models that rely exclusively on monolingual data have been shown to work \citep{DBLP:journals/corr/abs-1710-11041,DBLP:journals/corr/abs-1711-00043, artetxe-labaka-agirre:2018:EMNLP,lample2018phrase}.
Similarly, parallel data from other language pairs can be used to pre-train the network or jointly learn representations \citep{DBLP:journals/corr/ZophYMK16,chen-EtAl:2017:Long5,nguyen-chiang:2017:I17-2,neubig-hu:2018:EMNLP,gu-EtAl:2018:N18-1,gu-EtAl:2018:EMNLP1, kocmi-bojar:2018:WMT}.

While semi-supervised and unsupervised approaches have been shown to be very effective for some language pairs, their effectiveness depends on the availability of large amounts of suitable auxiliary data, and other conditions being met. For example, the effectiveness of unsupervised methods is impaired when languages are morphologically different, or when training domains do not match \cite{sgaard-ruder-vuli:2018:Long}

More broadly, this line of research still accepts the premise that NMT models are data-inefficient and require large amounts of auxiliary data to train.
In this work, we want to re-visit this point, and will focus on techniques to make more efficient use of small amounts of parallel training data.
Low-resource NMT without auxiliary data has received less attention; work in this direction includes \citep{DBLP:journals/corr/abs-1708-05729,nguyen-chiang:2018:N18-1}.

\section{Methods for Low-Resource Neural Machine Translation}

\label{sec:optimize}

\subsection{Mainstream Improvements}

We consider the hyperparameters used by \citet{koehn-knowles:2017:NMT} to be our baseline.
This baseline does not make use of various advances in NMT architectures and training tricks.
In contrast to the baseline, we use a BiDeep RNN architecture \citep{micelibarone2017}, label smoothing \citep{SzegedyEtAl2016}, dropout \citep{JMLR:v15:srivastava14a}, word dropout \citep{sennrich-haddow-birch:2016:WMT}, layer normalization \citep{DBLP:journals/corr/BaKH16} and tied embeddings \citep{DBLP:journals/corr/PressW16}.

\subsection{Language Representation}
\label{sec:bpe}

Subword representations such as BPE \citep{DBLP:journals/corr/SennrichHB15} have become a popular choice to achieve open-vocabulary translation.
BPE has one hyperparameter, the number of merge operations, which determines the size of the final vocabulary.
For high-resource settings, the effect of vocabulary size on translation quality is relatively small; \citet{haddow-EtAl:2018:WMT} report mixed results when comparing vocabularies of 30k and 90k subwords.

In low-resource settings, large vocabularies result in low-frequency (sub)words being represented as atomic units at training time, and the ability to learn good high-dimensional representations of these is doubtful.
\citet{uedin-nmt:2017} propose a minimum frequency threshold for subword units, and splitting any less frequent subword into smaller units or characters.
We expect that such a threshold reduces the need to carefully tune the vocabulary size to the dataset,
leading to more aggressive segmentation on smaller datasets.\footnote{In related work, \citet{2018arXiv180809943C} have shown that, given deep encoders and decoders, character-level models can outperform other subword segmentations.
In preliminary experiments, a character-level model performed poorly in our low-resource setting.}

\subsection{Hyperparameter Tuning}

Due to long training times, hyperparameters are hard to optimize by grid search, and are often re-used across experiments.
However, best practices differ between high-resource and low-resource settings.
While the trend in high-resource settings is towards using larger and deeper models, \citet{nguyen-chiang:2018:N18-1} use smaller and fewer layers for smaller datasets.
Previous work has argued for larger batch sizes in NMT \citep{morishita17nmt,neishi-EtAl:2017:WAT2017}, but we find that using smaller batches is beneficial in low-resource settings.
More aggressive dropout, including dropping whole words at random \citep{NIPS2016_6241}, is also likely to be more important.
We report results on a narrow hyperparameter search guided by previous work and our own intuition.

\subsection{Lexical Model}

Finally, we implement and test the lexical model by \citet{nguyen-chiang:2018:N18-1}, which has been shown to be beneficial in low-data conditions.
The core idea is to train a simple feed-forward network, the lexical model, jointly with the original attentional NMT model.
The input of the lexical model at time step $t$ is the weighted average of source embeddings $f$ (the attention weights $a$ are shared with the main model).
After a feedforward layer (with skip connection), the lexical model's output $h_t^l$ is combined with the original model's hidden state $h_t^o$ before softmax computation.
\begin{align*}
f_t^l &= \tanh \sum_s{a_t(s)f_s}\\
h_t^l &= \tanh(Wf_t^l) + f_t^l\\
p(y_t | y_{<t}, x) &= \text{softmax}(W^oh_t^o + b^o + W^lh_t^l + b^l)
\end{align*}
Our implementation adds dropout and layer normalization to the lexical model.\footnote{Implementation released in Nematus:\\ \url{https://github.com/EdinburghNLP/nematus}}

\section{Experiments}

\subsection{Data and Preprocessing}

We use the TED data from the IWSLT 2014 German$\to$English shared translation task \citep{iwslt2014}.
We use the same data cleanup and train/dev split as \citet{DBLP:journals/corr/RanzatoCAZ15}, resulting in \num{159000} parallel sentences of training data, and \num{7584} for development.

As a second language pair, we evaluate our systems on a Korean--English dataset\footnote{\url{https://sites.google.com/site/koreanparalleldata/}} with around \num{90000} parallel sentences of training data, \num{1000} for development, and \num{2000} for testing.

For both PBSMT and NMT, we apply the same tokenization and truecasing using Moses scripts.
For NMT, we also learn BPE subword segmentation with \num{30000} merge operations, shared between German and English, and independently for Korean$\to$English.

To simulate different amounts of training resources, we randomly subsample the IWSLT training corpus 5 times, discarding half of the data at each step.
Truecaser and BPE segmentation are learned on the full training corpus;
as one of our experiments, we set the frequency threshold for subword units to 10 in each subcorpus (see \ref{sec:bpe}).
Table \ref{datatbl} shows statistics for each subcorpus, including the subword vocabulary.

\begin{table}
\centering
\begin{tabular}{rrrr}
\toprule
& & \multicolumn{2}{c}{subword vocabulary}\\
\cmidrule(l){3-4}
sentences & words (EN) & \multicolumn{1}{c}{DE/KO} & \multicolumn{1}{c}{EN} \\
\midrule
DE$\to$EN\\
\num{159000} & \num{3220000} & \num{18870} & \num{13830} \\
\num{80000} &  \num{1610000} & \num{9850} & \num{7740} \\
\num{40000} &  \num{810000} & \num{7470} & \num{5950} \\
\num{20000} & \num{400000} & \num{5640} & \num{4530} \\
\num{10000} & \num{200000} & \num{3760} & \num{3110} \\
\num{5000} & \num{100000} & \num{2380} & \num{1990} \\
\midrule
KO$\to$EN\\
\num{94000} & \num{2300000} & \num{32082} & \num{16006} \\
\bottomrule
\end{tabular}
\caption{Training corpus size and subword vocabulary size for different subsets of IWSLT14 DE$\to$EN data, and for KO$\to$EN data.}
\label{datatbl}
\end{table}

Translation outputs are detruecased, detokenized, and compared against the reference with cased BLEU using sacreBLEU \citep{Papineni2002,post-2018-call}.\footnote{Signature BLEU+c.mixed+\#.1+s.exp+tok.13a+v.1.3.2.}
Like \citet{DBLP:journals/corr/RanzatoCAZ15}, we report BLEU on the concatenated dev sets for IWSLT 2014  (tst2010, tst2011, tst2012, dev2010, dev2012).

\subsection{PBSMT Baseline}

We use Moses \citep{koehnmoses} to train a PBSMT system.
We use MGIZA \citep{gao08} to train word alignments, and lmplz \citep{Heafield-estimate} for a 5-gram LM.
Feature weights are optimized on the dev set to maximize BLEU with batch MIRA \citep{Cherry:2012:BTS:2382029.2382089} -- we perform multiple runs where indicated.
Unlike \citet{koehn-knowles:2017:NMT}, we do not use extra data for the LM.
Both PBSMT and NMT can benefit from monolingual data, so the availability of monolingual data is no longer an exclusive advantage of PBSMT (see \ref{sec:transferlearning}).

\subsection{NMT Systems}

\begin{table*}
\centering
\begin{tabular}{clcc}
\toprule
& & \multicolumn{2}{c}{BLEU}\\
\cmidrule(l){3-4}
ID & system & 100k & 3.2M\\ 
\midrule
1 & phrase-based SMT & 15.87 $\pm$ 0.19 & 26.60 $\pm$ 0.00 \\ 
\midrule
2 & NMT baseline & \phantom{0}0.00 $\pm$ 0.00 & 25.70 $\pm$ 0.33 \\ 
\midrule
3 & 2 + "mainstream improvements" (dropout, tied embeddings, & \multirow{2}{*}{\phantom{0}7.20 $\pm$ 0.62}& \multirow{2}{*}{31.93 $\pm$ 0.05} \\ 
& \phantom{0 +} layer normalization, bideep RNN, label smoothing) & & \\
\midrule
4 & 3 + reduce BPE vocabulary (14k $\to$ 2k symbols) & 12.10 $\pm$ 0.16 & - \\ 
5 & 4 + reduce batch size (4k $\to$ 1k tokens) & 12.40 $\pm$ 0.08 & 31.97 $\pm$ 0.26 \\ 
6 & 5 + lexical model & 13.03 $\pm$ 0.49 & 31.80 $\pm$ 0.22 \\ 
\midrule
7 & 5 + aggressive (word) dropout & 15.87 $\pm$ 0.09 & \textbf{33.60} $\pm$ 0.14 \\ 
8 & 7 + other hyperparameter tuning (learning rate, & \multirow{2}{*}{\textbf{16.57} $\pm$ 0.26} & \multirow{2}{*}{32.80 $\pm$ 0.08}  \\ 
& \phantom{0 +} model depth, label smoothing rate) & & \\
9 & 8 + lexical model & 16.10 $\pm$ 0.29 & 33.30 $\pm$ 0.08 \\ 
\bottomrule
\end{tabular}
\caption{German$\to$English IWSLT results for training corpus size of 100k words and 3.2M words (full corpus). Mean and standard deviation of three training runs reported.}
\label{tbl:ablation}
\end{table*}

\begin{figure}
\centering
\begin{tikzpicture}[scale=0.8]
\pgfplotsset{major grid style={style=dotted}}
\pgfkeys{
    /pgf/number format/precision=1, 
    /pgf/number format/fixed=true, 
    /pgf/number format/fixed zerofill=false }
\begin{semilogxaxis}[xlabel=corpus size (English words),
    ymajorgrids=true,
    xmin=90000,
    xmax=4300000,
    ymin=0,
    ymax=36,
    ylabel=BLEU,
    legend pos = south east,
    legend style={
        font=\scriptsize,
        /tikz/nodes={anchor=west}
        },
    x tick label style={rotate=90},
    x label style={at={(axis description cs:0.5,-0.05)},anchor=north},
    mark size = 2,
    nodes near coords,
    every node near coord/.append style={font=\tiny,},
    ]

     \addplot +[rred, mark=x, raw gnuplot, solid, line width=0.2ex, id=nmt-optimizede, every node near coord/.append style={anchor=south,yshift=2pt}] gnuplot {plot 'plots/nmt-optimized.plot' using 2:6;};
     \addplot +[black, mark=o, raw gnuplot, solid, line width=0.2ex, id=smt, every node near coord/.append style={anchor=south}] gnuplot {plot 'plots/smt.plot' using 2:6;};
     \addplot +[ggreen, mark=square, raw gnuplot, solid, line width=0.2ex, id=nmt-baseline, every node near coord/.append style={anchor=north,xshift=8pt,yshift=2pt}] gnuplot {plot 'plots/nmt-baseline.plot' using 2:6;};

    \addlegendentry{neural MT optimized}
    \addlegendentry{phrase-based SMT}
    \addlegendentry{neural MT baseline}

\end{semilogxaxis}
\end{tikzpicture} 
\caption{German$\to$English learning curve, showing BLEU as a function of the amount of parallel training data, for PBSMT and NMT.}
\label{fg-learning}
\end{figure}
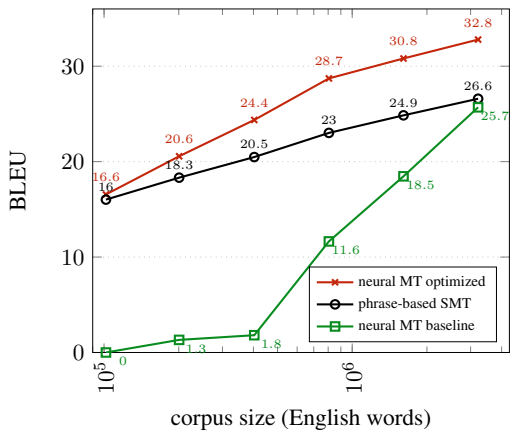

We train neural systems with Nematus \citep{nematus}.
Our baseline mostly follows the settings in \citep{koehn-knowles:2017:NMT}; we use adam \citep{DBLP:journals/corr/KingmaB14} and perform early stopping based on dev set BLEU.
We express our batch size in number of tokens, and set it to 4000 in the baseline (comparable to a batch size of 80 sentences used in previous work).

We subsequently add the methods described in section \ref{sec:optimize},
namely the bideep RNN, label smoothing, dropout, tied embeddings, layer normalization, changes to the BPE vocabulary size, batch size, model depth, regularization parameters and learning rate.
Detailed hyperparameters are reported in Appendix~\ref{sec:hyperparams}.

\begin{table*}
\centering
\begin{tabular}{lc}
\toprule
system & BLEU \\
\midrule
MIXER \citep{DBLP:journals/corr/RanzatoCAZ15}\tablefootnote{beam search results reported by \citet{DBLP:journals/corr/WisemanR16}.} & 21.8\phantom{0 $\pm$ 0.00} \\
BSO \citep{DBLP:journals/corr/WisemanR16} & 25.5\phantom{0 $\pm$ 0.00} \\
NPMT+LM \citep{huang2018towards} & 30.1\phantom{0 $\pm$ 0.00}\\
MRT \citep{edunov-etal-2018-classical} & 32.84 $\pm$ 0.08 \\
Pervasive Attention \citep{elbayad-besacier-verbeek:2018:K18-1} & 33.8\phantom{0 $\pm$ 0.00}\\
Transformer Baseline \citep{wu2018pay} & 34.4\phantom{0 $\pm$ 0.00} \\
Dynamic Convolution \citep{wu2018pay} & 35.2\phantom{0 $\pm$ 0.00} \\
\midrule
our PBSMT (1) & 28.19 $\pm$ 0.01 \\
our NMT baseline (2) & 27.16 $\pm$ 0.38 \\
our NMT best (7) & 35.27 $\pm$ 0.14 \\
\bottomrule
\end{tabular}
\caption{Results on full IWSLT14 German$\to$English data on tokenized and lowercased test set with \textit{multi-bleu.perl}.}
\label{lcbleu}
\end{table*}

\begin{table}
\centering
\begin{tabular}{lc}
\toprule
system & BLEU\\ 
\midrule
\cite{gu-EtAl:2018:EMNLP1} & \multirow{2}{*}{\phantom{0}5.97 \phantom{$\pm$ 0.00}}\\
(supervised Transformer) & \\
\midrule
phrase-based SMT & \phantom{0}6.57 $\pm$ 0.17 \\ 
NMT baseline (2) & \phantom{0}2.93 $\pm$ 0.34 \\ 
NMT optimized (8) & \textbf{10.37} $\pm$ 0.29 \\ 
\bottomrule
\end{tabular}
\caption{Korean$\to$English results. Mean and standard deviation of three training runs reported.}
\label{tbl:korean}
\end{table}

\section{Results}

Table \ref{tbl:ablation} shows the effect of adding different methods to the baseline NMT system,
on the ultra-low data condition (100k words of training data) and the full IWSLT 14 training corpus (3.2M words).
Our "mainstream improvements" add around 6--7 BLEU in both data conditions.

In the ultra-low data condition, reducing the BPE vocabulary size is very effective (+4.9 BLEU).
Reducing the batch size to 1000 token results in a BLEU gain of 0.3, and the lexical model yields an additional +0.6 BLEU.
However, aggressive (word) dropout\footnote{$p=0.3$ for dropping words; $p=0.5$ for other dropout.} (+3.4 BLEU) and tuning other hyperparameters (+0.7 BLEU) has a stronger effect than the lexical model, and adding the lexical model (9) on top of the optimized configuration (8) does not improve performance.
Together, the adaptations to the ultra-low data setting yield 9.4 BLEU (7.2$\to$16.6).
The model trained on full IWSLT data is less sensitive to our changes (31.9$\to$32.8 BLEU), and optimal hyperparameters differ depending on the data condition.
Subsequently, we still apply the hyperparameters that were optimized to the ultra-low data condition (8) to other data conditions, and Korean$\to$English, for simplicity.

For a comparison with PBSMT, and across different data settings, consider Figure \ref{fg-learning}, which shows the result of PBSMT, our NMT baseline, and our optimized NMT system.
Our NMT baseline still performs worse than the PBSMT system for 3.2M words of training data, which is consistent with the results by \citet{koehn-knowles:2017:NMT}.
However, our optimized NMT system shows strong improvements, and outperforms the PBSMT system across all data settings.
Some sample translations are shown in Appendix~\ref{sec:examples}.

For comparison to previous work, we report lowercased and tokenized results on the full IWSLT 14 training set in Table\ \ref{lcbleu}.
Our results far outperform the RNN-based results reported by \citet{DBLP:journals/corr/WisemanR16}, and are on par with the best reported results on this dataset.

Table \ref{tbl:korean} shows results for Korean$\to$English, using the same configurations (1, 2 and 8) as for German--English.
Our results confirm that the techniques we apply are successful across datasets, and result in stronger systems than previously reported on this dataset, achieving 10.37 BLEU as compared to 5.97 BLEU reported by \newcite{gu-EtAl:2018:EMNLP1}.

\section{Conclusions}

Our results demonstrate that NMT is in fact a suitable choice in low-data settings, and can outperform PBSMT with far less parallel training data than previously claimed.
Recently, the main trend in low-resource MT research has been the better exploitation of monolingual and multilingual resources.
Our results show that low-resource NMT is very sensitive to hyperparameters such as BPE vocabulary size, word dropout, and others, and by following a set of best practices, we can train competitive NMT systems without relying on auxiliary resources. This has practical relevance for languages where large amounts of monolingual data, or multilingual data involving related languages, are not available.
Even though we focused on only using parallel data, our results are also relevant for work on using auxiliary data to improve low-resource MT.
Supervised systems serve as an important baseline to judge the effectiveness of semisupervised or unsupervised approaches, and the quality of supervised systems trained on little data can directly impact semi-supervised workflows, for instance for the back-translation of monolingual data.

\section*{Acknowledgments}

Rico Sennrich has received funding from the Swiss National Science Foundation in the project CoNTra (grant number 105212\_169888).
Biao Zhang acknowledges the support of the Baidu Scholarship.

\bibliography{bibliography}

\begin{thebibliography}{53}
\expandafter\ifx\csname natexlab\endcsname\relax\def\natexlab#1{#1}\fi

\bibitem[{Artetxe et~al.(2018{\natexlab{a}})Artetxe, Labaka, and
  Agirre}]{artetxe-labaka-agirre:2018:EMNLP}
Mikel Artetxe, Gorka Labaka, and Eneko Agirre. 2018{\natexlab{a}}.
\newblock {Unsupervised Statistical Machine Translation}.
\newblock In \emph{{Proceedings of the 2018 Conference on Empirical Methods in
  Natural Language Processing}}, pages 3632--3642, Brussels, Belgium.

\bibitem[{Artetxe et~al.(2018{\natexlab{b}})Artetxe, Labaka, Agirre, and
  Cho}]{DBLP:journals/corr/abs-1710-11041}
Mikel Artetxe, Gorka Labaka, Eneko Agirre, and Kyunghyun Cho.
  2018{\natexlab{b}}.
\newblock {Unsupervised Neural Machine Translation}.
\newblock In \emph{{International Conference on Learning Representations}}.

\bibitem[{Ba et~al.(2016)Ba, Kiros, and Hinton}]{DBLP:journals/corr/BaKH16}
Lei~Jimmy Ba, Ryan Kiros, and Geoffrey~E. Hinton. 2016.
\newblock {Layer Normalization}.
\newblock \emph{CoRR}, abs/1607.06450.

\bibitem[{Bahdanau et~al.(2015)Bahdanau, Cho, and
  Bengio}]{DBLP:journals/corr/BahdanauCB14}
Dzmitry Bahdanau, Kyunghyun Cho, and Yoshua Bengio. 2015.
\newblock {Neural Machine Translation by Jointly Learning to Align and
  Translate}.
\newblock In \emph{{Proceedings of the International Conference on Learning
  Representations (ICLR)}}.

\bibitem[{Cettolo et~al.(2014)Cettolo, Niehues, St{\"u}ker, Bentivogli, and
  Federico}]{iwslt2014}
Mauro Cettolo, Jan Niehues, Sebastian St{\"u}ker, Luisa Bentivogli, and
  Marcello Federico. 2014.
\newblock {Report on the 11th IWSLT Evaluation Campaign, IWSLT 2014}.
\newblock In \emph{{Proceedings of the 11th Workshop on Spoken Language
  Translation}}, pages 2--16, Lake Tahoe, CA, USA.

\bibitem[{Chen et~al.(2017)Chen, Liu, Cheng, and Li}]{chen-EtAl:2017:Long5}
Yun Chen, Yang Liu, Yong Cheng, and Victor~O.K. Li. 2017.
\newblock {A Teacher-Student Framework for Zero-Resource Neural Machine
  Translation}.
\newblock In \emph{{Proceedings of the 55th Annual Meeting of the Association
  for Computational Linguistics (Volume 1: Long Papers)}}, pages 1925--1935,
  Vancouver, Canada.

\bibitem[{Cheng et~al.(2016)Cheng, Xu, He, He, Wu, Sun, and
  Liu}]{cheng-EtAl:2016:P16-1}
Yong Cheng, Wei Xu, Zhongjun He, Wei He, Hua Wu, Maosong Sun, and Yang Liu.
  2016.
\newblock {Semi-Supervised Learning for Neural Machine Translation}.
\newblock In \emph{{Proceedings of the 54th Annual Meeting of the Association
  for Computational Linguistics (Volume 1: Long Papers)}}, pages 1965--1974,
  Berlin, Germany.

\bibitem[{Cherry and Foster(2012)}]{Cherry:2012:BTS:2382029.2382089}
Colin Cherry and George Foster. 2012.
\newblock {Batch Tuning Strategies for Statistical Machine Translation}.
\newblock In \emph{{Proceedings of the 2012 Conference of the North American
  Chapter of the Association for Computational Linguistics: Human Language
  Technologies}}, {NAACL HLT '12}, pages 427--436, Montreal, Canada.

\bibitem[{Cherry et~al.(2018)Cherry, Foster, Bapna, Firat, and
  Macherey}]{2018arXiv180809943C}
Colin Cherry, George Foster, Ankur Bapna, Orhan Firat, and Wolfgang Macherey.
  2018.
\newblock {Revisiting Character-Based Neural Machine Translation with Capacity
  and Compression}.
\newblock In \emph{{Proceedings of the 2018 Conference on Empirical Methods in
  Natural Language Processing}}, pages 4295--4305, Brussels, Belgium.

\bibitem[{Currey et~al.(2017)Currey, {Miceli Barone}, and
  Heafield}]{currey-micelibarone-heafield:2017:WMT}
Anna Currey, Antonio~Valerio {Miceli Barone}, and Kenneth Heafield. 2017.
\newblock {Copied Monolingual Data Improves Low-Resource Neural Machine
  Translation}.
\newblock In \emph{{Proceedings of the Second Conference on Machine
  Translation}}, pages 148--156, Copenhagen, Denmark.

\bibitem[{Edunov et~al.(2018)Edunov, Ott, Auli, Grangier, and
  Ranzato}]{edunov-etal-2018-classical}
Sergey Edunov, Myle Ott, Michael Auli, David Grangier, and Marc{'}Aurelio
  Ranzato. 2018.
\newblock Classical structured prediction losses for sequence to sequence
  learning.
\newblock In \emph{Proceedings of the 2018 Conference of the North {A}merican
  Chapter of the Association for Computational Linguistics: Human Language
  Technologies, Volume 1 (Long Papers)}, pages 355--364, New Orleans,
  Louisiana.

\bibitem[{Elbayad et~al.(2018)Elbayad, Besacier, and
  Verbeek}]{elbayad-besacier-verbeek:2018:K18-1}
Maha Elbayad, Laurent Besacier, and Jakob Verbeek. 2018.
\newblock {Pervasive Attention: {2D} Convolutional Neural Networks for
  Sequence-to-Sequence Prediction}.
\newblock In \emph{{Proceedings of the 22nd Conference on Computational Natural
  Language Learning}}, pages 97--107, Brussels, Belgium.

\bibitem[{Gal and Ghahramani(2016)}]{NIPS2016_6241}
Yarin Gal and Zoubin Ghahramani. 2016.
\newblock A theoretically grounded application of dropout in recurrent neural
  networks.
\newblock In \emph{Advances in Neural Information Processing Systems 29}, pages
  1019--1027.

\bibitem[{Gao and Vogel(2008)}]{gao08}
Qin Gao and Stephan Vogel. 2008.
\newblock {Parallel Implementations of Word Alignment Tool}.
\newblock In \emph{{Software Engineering, Testing, and Quality Assurance for
  Natural Language Processing}}, pages 49--57, Columbus, Ohio.

\bibitem[{Gu et~al.(2018{\natexlab{a}})Gu, Hassan, Devlin, and
  Li}]{gu-EtAl:2018:N18-1}
Jiatao Gu, Hany Hassan, Jacob Devlin, and Victor~O.K. Li. 2018{\natexlab{a}}.
\newblock {Universal Neural Machine Translation for Extremely Low Resource
  Languages}.
\newblock In \emph{{Proceedings of the 2018 Conference of the North American
  Chapter of the Association for Computational Linguistics: Human Language
  Technologies, Volume 1 (Long Papers)}}, pages 344--354, New Orleans,
  Louisiana.

\bibitem[{Gu et~al.(2018{\natexlab{b}})Gu, Wang, Chen, Li, and
  Cho}]{gu-EtAl:2018:EMNLP1}
Jiatao Gu, Yong Wang, Yun Chen, Victor O.~K. Li, and Kyunghyun Cho.
  2018{\natexlab{b}}.
\newblock {Meta-Learning for Low-Resource Neural Machine Translation}.
\newblock In \emph{{Proceedings of the 2018 Conference on Empirical Methods in
  Natural Language Processing}}, pages 3622--3631, Brussels, Belgium.

\bibitem[{G{\"u}l{\c c}ehre et~al.(2015)G{\"u}l{\c c}ehre, Firat, Xu, Cho,
  Barrault, Lin, Bougares, Schwenk, and
  Bengio}]{DBLP:journals/corr/GulcehreFXCBLBS15}
{\c C}aglar G{\"u}l{\c c}ehre, Orhan Firat, Kelvin Xu, Kyunghyun Cho,
  {Lo\"{\i}c} Barrault, Huei{-}Chi Lin, Fethi Bougares, Holger Schwenk, and
  Yoshua Bengio. 2015.
\newblock {On Using Monolingual Corpora in Neural Machine Translation}.
\newblock \emph{CoRR}, abs/1503.03535.

\bibitem[{Haddow et~al.(2018)Haddow, Bogoychev, Emelin, Germann, Grundkiewicz,
  Heafield, {Miceli Barone}, and Sennrich}]{haddow-EtAl:2018:WMT}
Barry Haddow, Nikolay Bogoychev, Denis Emelin, Ulrich Germann, Roman
  Grundkiewicz, Kenneth Heafield, Antonio~Valerio {Miceli Barone}, and Rico
  Sennrich. 2018.
\newblock {The University of Edinburgh{\rq}s Submissions to the WMT18 News
  Translation Task}.
\newblock In \emph{{Proceedings of the Third Conference on Machine
  Translation}}, pages 403--413, Belgium, Brussels.

\bibitem[{He et~al.(2016)He, Xia, Qin, Wang, Yu, Liu, and Ma}]{NIPS2016_6469}
Di~He, Yingce Xia, Tao Qin, Liwei Wang, Nenghai Yu, Tieyan Liu, and Wei-Ying
  Ma. 2016.
\newblock {Dual Learning for Machine Translation}.
\newblock In \emph{{Advances in Neural Information Processing Systems 29}},
  pages 820--828.

\bibitem[{Heafield et~al.(2013)Heafield, Pouzyrevsky, Clark, and
  Koehn}]{Heafield-estimate}
Kenneth Heafield, Ivan Pouzyrevsky, Jonathan~H. Clark, and Philipp Koehn. 2013.
\newblock {Scalable Modified {Kneser-Ney} Language Model Estimation}.
\newblock In \emph{{Proceedings of the 51st Annual Meeting of the Association
  for Computational Linguistics}}, pages 690--696, Sofia, Bulgaria.

\bibitem[{Huang et~al.(2018)Huang, Wang, Huang, Zhou, and
  Deng}]{huang2018towards}
Po-Sen Huang, Chong Wang, Sitao Huang, Dengyong Zhou, and Li~Deng. 2018.
\newblock {Towards Neural Phrase-based Machine Translation}.
\newblock In \emph{{International Conference on Learning Representations}}.

\bibitem[{Kingma and Ba(2015)}]{DBLP:journals/corr/KingmaB14}
Diederik~P. Kingma and Jimmy Ba. 2015.
\newblock {Adam: {A} Method for Stochastic Optimization}.
\newblock In \emph{{The International Conference on Learning Representations}},
  San Diego, California, USA.

\bibitem[{Kocmi and Bojar(2018)}]{kocmi-bojar:2018:WMT}
Tom Kocmi and Ond{\v r}ej Bojar. 2018.
\newblock {Trivial Transfer Learning for Low-Resource Neural Machine
  Translation}.
\newblock In \emph{{Proceedings of the Third Conference on Machine
  Translation}}, pages 244--252, Belgium, Brussels.

\bibitem[{Koehn et~al.(2007)Koehn, Hoang, Birch, Callison-Burch, Federico,
  Bertoldi, Cowan, Shen, Moran, Zens, Dyer, Bojar, Constantin, and
  Herbst}]{koehnmoses}
Philipp Koehn, Hieu Hoang, Alexandra Birch, Chris Callison-Burch, Marcello
  Federico, Nicola Bertoldi, Brooke Cowan, Wade Shen, Christine Moran, Richard
  Zens, Chris Dyer, Ond{\v r}ej Bojar, Alexandra Constantin, and Evan Herbst.
  2007.
\newblock {Moses: Open Source Toolkit for Statistical Machine Translation}.
\newblock In \emph{{Proceedings of the ACL-2007 Demo and Poster Sessions}},
  pages 177--180, Prague, Czech Republic.

\bibitem[{Koehn and Knowles(2017)}]{koehn-knowles:2017:NMT}
Philipp Koehn and Rebecca Knowles. 2017.
\newblock {Six Challenges for Neural Machine Translation}.
\newblock In \emph{{Proceedings of the First Workshop on Neural Machine
  Translation}}, pages 28--39, Vancouver.

\bibitem[{Lample et~al.(2018{\natexlab{a}})Lample, Denoyer, and
  Ranzato}]{DBLP:journals/corr/abs-1711-00043}
Guillaume Lample, Ludovic Denoyer, and Marc'Aurelio Ranzato.
  2018{\natexlab{a}}.
\newblock {Unsupervised Machine Translation Using Monolingual Corpora Only}.
\newblock In \emph{{International Conference on Learning Representations}}.

\bibitem[{Lample et~al.(2018{\natexlab{b}})Lample, Ott, Conneau, Denoyer, and
  Ranzato}]{lample2018phrase}
Guillaume Lample, Myle Ott, Alexis Conneau, Ludovic Denoyer, and Marc'Aurelio
  Ranzato. 2018{\natexlab{b}}.
\newblock {Phrase-Based \& Neural Unsupervised Machine Translation}.
\newblock In \emph{{Proceedings of the 2018 Conference on Empirical Methods in
  Natural Language Processing (EMNLP)}}, pages 5039--5049, Brussels, Belgium.

\bibitem[{Luong et~al.(2016)Luong, Le, Sutskever, Vinyals, and
  Kaiser}]{DBLP:journals/corr/LuongLSVK15}
Minh{-}Thang Luong, Quoc~V. Le, Ilya Sutskever, Oriol Vinyals, and Lukasz
  Kaiser. 2016.
\newblock {Multi-task Sequence to Sequence Learning}.
\newblock In \emph{{The International Conference on Learning Representations}}.

\bibitem[{{Miceli Barone} et~al.(2017){Miceli Barone}, Helcl, Sennrich, Haddow,
  and Birch}]{micelibarone2017}
Antonio~Valerio {Miceli Barone}, Jind{\v r}ich Helcl, Rico Sennrich, Barry
  Haddow, and Alexandra Birch. 2017.
\newblock {Deep Architectures for Neural Machine Translation}.
\newblock In \emph{{Proceedings of the Second Conference on Machine
  Translation, Volume 1: Research Papers}}, Copenhagen, Denmark.

\bibitem[{Morishita et~al.(2017)Morishita, Oda, Neubig, Yoshino, Sudoh, and
  Nakamura}]{morishita17nmt}
Makoto Morishita, Yusuke Oda, Graham Neubig, Koichiro Yoshino, Katsuhito Sudoh,
  and Satoshi Nakamura. 2017.
\newblock {An Empirical Study of Mini-Batch Creation Strategies for Neural
  Machine Translation}.
\newblock In \emph{{The First Workshop on Neural Machine Translation (NMT)}},
  pages 61--68, Vancouver, Canada.

\bibitem[{Neishi et~al.(2017)Neishi, Sakuma, Tohda, Ishiwatari, Yoshinaga, and
  Toyoda}]{neishi-EtAl:2017:WAT2017}
Masato Neishi, Jin Sakuma, Satoshi Tohda, Shonosuke Ishiwatari, Naoki
  Yoshinaga, and Masashi Toyoda. 2017.
\newblock {A Bag of Useful Tricks for Practical Neural Machine Translation:
  Embedding Layer Initialization and Large Batch Size}.
\newblock In \emph{{Proceedings of the 4th Workshop on Asian Translation
  (WAT2017)}}, pages 99--109, Taipei, Taiwan.

\bibitem[{Neubig and Hu(2018)}]{neubig-hu:2018:EMNLP}
Graham Neubig and Junjie Hu. 2018.
\newblock {Rapid Adaptation of Neural Machine Translation to New Languages}.
\newblock In \emph{{Proceedings of the 2018 Conference on Empirical Methods in
  Natural Language Processing}}, pages 875--880, Brussels, Belgium.

\bibitem[{Nguyen and Chiang(2018)}]{nguyen-chiang:2018:N18-1}
Toan Nguyen and David Chiang. 2018.
\newblock {Improving Lexical Choice in Neural Machine Translation}.
\newblock In \emph{{Proceedings of the 2018 Conference of the North American
  Chapter of the Association for Computational Linguistics: Human Language
  Technologies, Volume 1 (Long Papers)}}, pages 334--343, New Orleans,
  Louisiana.

\bibitem[{Nguyen and Chiang(2017)}]{nguyen-chiang:2017:I17-2}
Toan~Q. Nguyen and David Chiang. 2017.
\newblock {Transfer Learning across Low-Resource, Related Languages for Neural
  Machine Translation}.
\newblock In \emph{{Proceedings of the Eighth International Joint Conference on
  Natural Language Processing (Volume 2: Short Papers)}}, pages 296--301,
  Taipei, Taiwan.

\bibitem[{{\"O}stling and Tiedemann(2017)}]{DBLP:journals/corr/abs-1708-05729}
Robert {\"O}stling and J{\"o}rg Tiedemann. 2017.
\newblock {Neural machine translation for low-resource languages}.
\newblock \emph{CoRR}, abs/1708.05729.

\bibitem[{Papineni et~al.(2002)Papineni, Roukos, Ward, and Zhu}]{Papineni2002}
Kishore Papineni, Salim Roukos, Todd Ward, and Wei-Jing Zhu. 2002.
\newblock {{BLEU}: A Method for Automatic Evaluation of Machine Translation}.
\newblock In \emph{{Proceedings of the 40th Annual Meeting on Association for
  Computational Linguistics}}, pages 311--318, Philadelphia, PA.

\bibitem[{Post(2018)}]{post-2018-call}
Matt Post. 2018.
\newblock A call for clarity in reporting {BLEU} scores.
\newblock In \emph{Proceedings of the Third Conference on Machine Translation:
  Research Papers}, pages 186--191, Belgium, Brussels.

\bibitem[{Press and Wolf(2017)}]{DBLP:journals/corr/PressW16}
Ofir Press and Lior Wolf. 2017.
\newblock {Using the Output Embedding to Improve Language Models}.
\newblock In \emph{{Proceedings of the 15th Conference of the European Chapter
  of the Association for Computational Linguistics (EACL)}}, Valencia, Spain.

\bibitem[{Ramachandran et~al.(2017)Ramachandran, Liu, and
  Le}]{ramachandran2016unsupervised}
Prajit Ramachandran, Peter Liu, and Quoc Le. 2017.
\newblock {Unsupervised pretraining for sequence to sequence learning}.
\newblock In \emph{{Proceedings of the 2017 Conference on Empirical Methods in
  Natural Language Processing}}, pages 383--391, Copenhagen, Denmark.

\bibitem[{Ranzato et~al.(2016)Ranzato, Chopra, Auli, and
  Zaremba}]{DBLP:journals/corr/RanzatoCAZ15}
Marc'Aurelio Ranzato, Sumit Chopra, Michael Auli, and Wojciech Zaremba. 2016.
\newblock {Sequence Level Training with Recurrent Neural Networks}.
\newblock In \emph{{The International Conference on Learning Representations}}.

\bibitem[{Sennrich et~al.(2017{\natexlab{a}})Sennrich, Birch, Currey, Germann,
  Haddow, Heafield, {Miceli Barone}, and Williams}]{uedin-nmt:2017}
Rico Sennrich, Alexandra Birch, Anna Currey, Ulrich Germann, Barry Haddow,
  Kenneth Heafield, Antonio~Valerio {Miceli Barone}, and Philip Williams.
  2017{\natexlab{a}}.
\newblock {The University of Edinburgh's Neural MT Systems for WMT17}.
\newblock In \emph{{Proceedings of the Second Conference on Machine
  Translation, Volume 2: Shared Task Papers}}, Copenhagen, Denmark.

\bibitem[{Sennrich et~al.(2017{\natexlab{b}})Sennrich, Firat, Cho, Birch,
  Haddow, Hitschler, Junczys-Dowmunt, L{\"a}ubli, {Miceli Barone}, Mokry, and
  Nadejde}]{nematus}
Rico Sennrich, Orhan Firat, Kyunghyun Cho, Alexandra Birch, Barry Haddow,
  Julian Hitschler, Marcin Junczys-Dowmunt, Samuel L{\"a}ubli, Antonio~Valerio
  {Miceli Barone}, Jozef Mokry, and Maria Nadejde. 2017{\natexlab{b}}.
\newblock {Nematus: a Toolkit for Neural Machine Translation}.
\newblock In \emph{{Proceedings of the Software Demonstrations of the 15th
  Conference of the European Chapter of the Association for Computational
  Linguistics}}, pages 65--68, Valencia, Spain.

\bibitem[{Sennrich et~al.(2016{\natexlab{a}})Sennrich, Haddow, and
  Birch}]{sennrich-haddow-birch:2016:WMT}
Rico Sennrich, Barry Haddow, and Alexandra Birch. 2016{\natexlab{a}}.
\newblock {Edinburgh Neural Machine Translation Systems for WMT 16}.
\newblock In \emph{{Proceedings of the First Conference on Machine Translation,
  Volume 2: Shared Task Papers}}, pages 368--373, Berlin, Germany.

\bibitem[{Sennrich et~al.(2016{\natexlab{b}})Sennrich, Haddow, and
  Birch}]{2015arXiv151106709S}
Rico Sennrich, Barry Haddow, and Alexandra Birch. 2016{\natexlab{b}}.
\newblock {Improving Neural Machine Translation Models with Monolingual Data}.
\newblock In \emph{{Proceedings of the 54th Annual Meeting of the Association
  for Computational Linguistics (Volume 1: Long Papers)}}, pages 86--96,
  Berlin, Germany.

\bibitem[{Sennrich et~al.(2016{\natexlab{c}})Sennrich, Haddow, and
  Birch}]{DBLP:journals/corr/SennrichHB15}
Rico Sennrich, Barry Haddow, and Alexandra Birch. 2016{\natexlab{c}}.
\newblock {Neural Machine Translation of Rare Words with Subword Units}.
\newblock In \emph{{Proceedings of the 54th Annual Meeting of the Association
  for Computational Linguistics (Volume 1: Long Papers)}}, pages 1715--1725,
  Berlin, Germany.

\bibitem[{S{\o}gaard et~al.(2018)S{\o}gaard, Ruder, and
  Vulic}]{sgaard-ruder-vuli:2018:Long}
Anders S{\o}gaard, Sebastian Ruder, and Ivan Vulic. 2018.
\newblock On the limitations of unsupervised bilingual dictionary induction.
\newblock In \emph{Proceedings of the 56th Annual Meeting of the Association
  for Computational Linguistics (Volume 1: Long Papers)}, pages 778--788,
  Melbourne, Australia.

\bibitem[{Srivastava et~al.(2014)Srivastava, Hinton, Krizhevsky, Sutskever, and
  Salakhutdinov}]{JMLR:v15:srivastava14a}
Nitish Srivastava, Geoffrey Hinton, Alex Krizhevsky, Ilya Sutskever, and Ruslan
  Salakhutdinov. 2014.
\newblock {Dropout: A Simple Way to Prevent Neural Networks from Overfitting}.
\newblock \emph{Journal of Machine Learning Research}, 15:1929--1958.

\bibitem[{Sutskever et~al.(2014)Sutskever, Vinyals, and
  Le}]{DBLP:conf/nips/SutskeverVL14}
Ilya Sutskever, Oriol Vinyals, and Quoc~V. Le. 2014.
\newblock {Sequence to Sequence Learning with Neural Networks}.
\newblock In \emph{{Advances in Neural Information Processing Systems 27:
  Annual Conference on Neural Information Processing Systems 2014}}, pages
  3104--3112, Montreal, Quebec, Canada.

\bibitem[{Szegedy et~al.(2016)Szegedy, Vanhoucke, Ioffe, Shlens, and
  Wojna}]{SzegedyEtAl2016}
Christian Szegedy, Vincent Vanhoucke, Sergey Ioffe, Jonathon Shlens, and
  Z.~Wojna. 2016.
\newblock {Rethinking the Inception Architecture for Computer Vision}.
\newblock In \emph{{2016 IEEE Conference on Computer Vision and Pattern
  Recognition (CVPR)}}, pages 2818--2826.

\bibitem[{Vaswani et~al.(2017)Vaswani, Shazeer, Parmar, Uszkoreit, Jones,
  Gomez, Kaiser, and Polosukhin}]{DBLP:journals/corr/VaswaniSPUJGKP17}
Ashish Vaswani, Noam Shazeer, Niki Parmar, Jakob Uszkoreit, Llion Jones,
  Aidan~N Gomez, {\L}ukasz Kaiser, and Illia Polosukhin. 2017.
\newblock {Attention is All you Need}.
\newblock In \emph{{Advances in Neural Information Processing Systems 30}},
  pages 5998--6008.

\bibitem[{Wiseman and Rush(2016)}]{DBLP:journals/corr/WisemanR16}
Sam Wiseman and Alexander~M. Rush. 2016.
\newblock Sequence-to-sequence learning as beam-search optimization.
\newblock In \emph{Proceedings of the 2016 Conference on Empirical Methods in
  Natural Language Processing}, pages 1296--1306, Austin, Texas.

\bibitem[{Wu et~al.(2019)Wu, Fan, Baevski, Dauphin, and Auli}]{wu2018pay}
Felix Wu, Angela Fan, Alexei Baevski, Yann Dauphin, and Michael Auli. 2019.
\newblock Pay less attention with lightweight and dynamic convolutions.
\newblock In \emph{International Conference on Learning Representations}.

\bibitem[{Zoph et~al.(2016)Zoph, Yuret, May, and
  Knight}]{DBLP:journals/corr/ZophYMK16}
Barret Zoph, Deniz Yuret, Jonathan May, and Kevin Knight. 2016.
\newblock {Transfer Learning for Low-Resource Neural Machine Translation}.
\newblock In \emph{{Proceedings of the 2016 Conference on Empirical Methods in
  Natural Language Processing}}, pages 1568--1575, Austin, Texas.

\end{thebibliography}
\bibliographystyle{acl_natbib}

\newpage
\appendix

\section{Hyperparameters}
\label{sec:hyperparams}

Table \ref{tab:params} lists hyperparameters used for the different experiments in the ablation study (Table 2). Hyperparameters were kept constant across different data settings, except for the validation interval and subword vocabulary size (see Table 1).

\begin{table*}
\centering
\begin{tabular}{lccccccc}
\toprule
& \multicolumn{7}{c}{system}\\
\cmidrule(l){2-8}
hyperparameter & 2 & 3 & 5 & 6 & 7 & 8 & 9\\
\midrule
hidden layer size & 1024\\
embedding size & 512 \\
encoder depth & 1 & 2 & & & & 1\\
encoder recurrence transition depth & 1 & 2 \\
decoder depth & 1 & 2 & & & & 1\\
dec.\ recurrence transition depth (base) & 2 & 4 & & & & 2\\
dec.\ recurrence transition depth (high) & - & 2 & & & & -\\
tie decoder embeddings & - & yes \\
layer normalization & - & yes \\
lexical model & - & & & yes & - & & yes\\
\midrule
hidden dropout & - & 0.2 &  & & 0.5\\
embedding dropout & - & 0.2 &  & & 0.5\\
source word dropout & - & 0.1 &  & & 0.3\\
target word dropout & - &  &  & & 0.3\\
label smoothing & - & 0.1 & & & & 0.2\\
\midrule
maximum sentence length & 200 \\
minibatch size (\# tokens) & 4000 & & 1000 & \\
learning rate & 0.0001 & & & & & 0.0005 \\
optimizer & adam & & & \\
early stopping patience & 10\\
validation interval: & \\
\hspace{0.5cm} IWSLT 100k / 200k / 400k & 50 & 100 & 400 & \\
\hspace{0.5cm} IWSLT $\geq$ 800k / KO-EN 2.3M & 1000 & 2000 & 8000 & \\
\midrule
beam size & 5\\
\bottomrule
\end{tabular}
\caption{Configurations of NMT systems reported in Table 2. Empty fields indicate that hyperparameter was unchanged compared to previous systems.}
\label{tab:params}
\end{table*}

\section{Sample Translations}
\label{sec:examples}

\begin{table*}
\centering
\small

\begin{tabular}{lp{12cm}} 
\toprule
source & In einem blutbefleckten Kontinent, waren diese Menschen die einzigen, die nie von den Spaniern erobert wurden.\\
reference & In a bloodstained continent, these people alone were never conquered by the Spanish.\\
\midrule
PBSMT 100k & In a blutbefleckten continent, were these people the only, the never of the Spaniern erobert were.\\
PBSMT 3.2M & In a blutbefleckten continent, these people were the only ones that were never of the Spaniern conquered.\\
\midrule
NMT 3.2M (baseline) & In a blinging tree continent, these people were the only ones that never had been conquered by the Spanians.\\
\midrule
NMT 100k (optimized) & In a blue-flect continent, these people were the only one that has never been doing by the spaniers.\\
NMT 3.2M (optimized) & In a bleed continent, these people were the only ones who had never been conquered by the Spanians.\\
\addlinespace[10pt]
\toprule 
\addlinespace[10pt]
source & Dies ist tatsächlich ein Poster von Notre Dame, das richtig aufgezeichnet wurde. \\
reference & This is actually a poster of Notre Dame that registered correctly.\\
\midrule
PBSMT 100k & This is actually poster of Notre lady, the right aufgezeichnet was.\\
PBSMT 3.2M & This is actually a poster of Notre Dame, the right recorded. \\
\midrule
NMT 3.2M (baseline) & This is actually a poster of emergency lady who was just recorded properly.\\
\midrule
NMT 100k (optimized) & This is actually a poster of Notre Dame, that was really the first thing.\\
NMT 3.2M (optimized) & This is actually a poster from Notre Dame, which has been recorded right.\\
\bottomrule
\end{tabular}
\caption{German$\to$English translation examples with phrase-based SMT and NMT systems trained on 100k/3.2M words of parallel data.}
\label{transl}
\end{table*}

Table \ref{transl} shows some sample translations that represent typical errors of our PBSMT and NMT systems, trained with ultra-low (100k words) and low (3.2M words) amounts of data.
For unknown words such as \emph{blutbefleckten} (`bloodstained') or \emph{Spaniern} (`Spaniards', `Spanish'), PBSMT systems default to copying,
while NMT systems produce translations on a subword-level, with varying success (\emph{blue-flect}, \emph{bleed}; \emph{spaniers}, \emph{Spanians}).
NMT systems learn some syntactic disambiguation even with very little data, for example the translation of \emph{das} and \emph{die} as relative pronouns ('that', 'which', 'who'), while PBSMT produces less grammatical translation.
On the flip side, the ultra low-resource NMT system ignores some unknown words in favour of a more-or-less fluent, but semantically inadequate translation: \emph{erobert} ('conquered') is translated into \emph{doing}, and \emph{richtig aufgezeichnet} ('registered correctly', `recorded correctly') into \emph{really the first thing}.

\end{document}